\PassOptionsToPackage{expansion=false}{microtype}
\documentclass[sigconf,10pt,natbib,nonacm]{acmart}

\renewcommand\footnotetextcopyrightpermission[1]{}
\acmConference{}%
\acmYear{}%
\acmISBN{}%
\acmDOI{}%
\setcopyright{none}

\usepackage{amsmath}
\usepackage{booktabs}
\usepackage{array}
\usepackage{multirow}
\usepackage{subcaption}
\usepackage{float}
\usepackage{enumitem}
\usepackage{pifont}
\usepackage{graphicx}
\usepackage{placeins}
\usepackage[most]{tcolorbox}
\usepackage{tikz}
\usepackage{pgfplots}
\pgfplotsset{compat=1.17}
\usepackage{url}

\ifxetex
  \newfontfamily{\bengalifont}{NotoSansBengali}[
    Path=./,
    Extension=.ttf,
    UprightFont=*-Regular,
    Script=Bengali
  ]
  
\else
  
\fi

\newcommand{\cmark}{\ding{51}}
\newcommand{\xmark}{\ding{55}}



\title{Where Does Retrieval Fail? Evaluating RAG Architectures for Agricultural Advisory}

\author{Khan Raiyan Ibne Reza \texorpdfstring{\quad}{ } Sanjana Aktar Maria \texorpdfstring{\quad}{ } Sumaiya Tabassum Nimi}
\affiliation{%
  \institution{North South University}
  \city{Dhaka}
  \country{Bangladesh}
}
\email{{raiyan.reza, sanjana.maria, sumaiya.nimi}@northsouth.edu}

\begin{document}

\begin{abstract}
Retrieval quality in RAG systems is commonly reported as a single aggregate score, which can hide large differences across query types and language conditions. We study this problem in Bengali agricultural advisory, where farmer queries are often colloquial while official advisory documents use formal scientific terminology. We construct a test collection of 1,000 queries and 2,882 knowledge nodes extracted from 284 official Bangladeshi agricultural publications, and use it to evaluate five retrieval architectures and six embedding models under three controlled language conditions.

The results show that no single retrieval method is consistently best. For native Bengali queries, BM25 is the strongest single retriever (R@10${=}0.506$) while Hybrid RRF reaches the highest overall R@10 of 0.539. However, dense retrieval performance varies sharply by query type: R@10 is 0.093 on colloquial farmer queries and 0.970 on formal safety queries. Across language conditions, BM25 R@10 drops from 0.506 on Bengali queries to 0.004 when English queries are matched against the Bengali corpus, while dense retrieval falls only from 0.464 to 0.425. We also find that embedding task configuration and passage length can each change reported R@10 by a factor of seven, independent of architecture. These results show why low-resource RAG evaluation should report performance by language condition and query type rather than relying on aggregate scores alone. The dataset and evaluation scripts are available at \url{https://huggingface.co/datasets/RaiyanKhaan/AgriTrust-RAG}.
\end{abstract}

\maketitle

\section{Introduction}

\subsection{Motivation and Gap}
Retrieval-Augmented Generation (RAG) systems are increasingly deployed in domain-specific settings, yet evaluated almost exclusively on English benchmarks. Bengali agricultural advisory is a useful test case: with 237 million speakers~\cite{eberhard2024ethnologue}, advisory queries use colloquial farmer language while authoritative source documents use formal scientific Bengali. This register gap challenges lexical and dense retrieval in opposing ways, and the stakes are concrete: 200 of the benchmark's 1,000 queries cover safety-critical content (pesticide dosage, off-label chemical use), where a retrieval failure has real-world consequences.

Existing retrieval benchmarks (BEIR~\cite{thakur2021beir}, CRAG~\cite{yang2024crag}) are English-only; multilingual collections like MIRACL~\cite{zhang2023miracl} and Mr.\ TyDi~\cite{zhang2021mr} do not target agricultural advisory or query-register variation. Prior Bengali RAG systems translate queries before retrieval~\cite{hossain2026cost} or do not isolate the retrieval layer~\cite{nawal2024effective}, while the closest agricultural resource~\cite{reza2026krishokchat} is not a retrieval-only benchmark with provenance-level gold labels or controlled cross-lingual conditions.

\subsection{The Benchmark Collection}
Unlike existing retrieval benchmarks that pair text chunks with aggregate scores, our benchmark is designed around six properties that form its benchmark design: traceable retrieval units, structured canonical nodes rather than raw text, full document traceability, a controlled multilingual protocol, agricultural domain grounding, and register-aware evaluation covering both farmer queries and formal queries. The benchmark consists of: (i) 284 source PDFs from five Bangladeshi government and research organizations; (ii) 2,882 knowledge nodes (1,022 image-linked), each a traceable retrieval unit; (iii) a knowledge graph of 19,768 entities and 17,501 factual triples; (iv) 1,000 queries (900 answerable) across three categories (farmer-anchored, KG-grounded, safety), with gold-node mappings verified by three independent annotators (Fleiss' $\kappa{=}0.72$); and (v) an evaluation harness implementing strict document-level gold matching, bootstrap confidence intervals, and an embedding API configuration audit.

Evaluating on this collection reveals four key findings, summarized in the abstract and developed in full in Section~\ref{sec:results}, with analysis of these patterns in Section~\ref{sec:mechanisms}.

\subsection{Contributions}
\begin{enumerate}[leftmargin=*, label=\arabic*., itemsep=0pt, topsep=1pt]
  \item \textbf{Benchmarking collection.} A rigorous test collection for low-resource Bengali agricultural retrieval, covering 284 authoritative documents, 2,882 knowledge nodes, 19,768 entities, 17,501 triples, and 1,000 annotated queries (available at \url{https://huggingface.co/datasets/RaiyanKhaan/AgriTrust-RAG}).
  \item \textbf{Canonical node representation.} A layered retrieval unit uniting natural-language content, structured facts, and deterministic source metadata via bounded, schema-guided extraction.
  \item \textbf{Evaluation harness \& audit protocol.} A reproducible framework implementing strict gold matching, bootstrap confidence intervals, and an embedding API configuration audit.
  \item \textbf{Benchmark validation via systematic diagnosis.} Evaluation across five architectures, six embedding models, and three language conditions demonstrates that the benchmark's register- and script-stratified design surfaces failure modes invisible to aggregate, single-condition evaluation.
\end{enumerate}

\section{Related Work}
\label{sec:related}

\subsection{IR and RAG Evaluation Benchmarks}
Standard RAG evaluation benchmarks such as BEIR~\cite{thakur2021beir}, RAGBench~\cite{friel2024ragbench}, GaRAGe~\cite{sorodoc2025garage}, and CRAG~\cite{yang2024crag} are built almost entirely in English; recent extensions such as T2-RAGBench~\cite{strich2026t2} target text-and-table retrieval in financial documents, not the provenance-grounded, low-resource domain advisory setting we address. Multilingual retrieval benchmarks such as MIRACL~\cite{zhang2023miracl} (18 languages) and Mr.\ TyDi~\cite{zhang2021mr} extend evaluation to diverse languages, but none enable controlled architecture-isolated retrieval evaluation with provenance grounding in a low-resource agricultural setting. Standard benchmarks typically emphasize aggregate accuracy scores, concealing subgroup failure across query registers, a limitation highlighted by concurrent work on stratified retrieval evaluation~\cite{klearman2026coverage}.

\subsection{Retrieval in Low-Resource Languages}
Dense retrieval often degrades outside high-resource languages due to tokenization fragmentation and language bias~\cite{hong2026improving, lee2026clear, yeshambel2025dense}. Cross-lingual DPR shows modest gains~\cite{wu2024limits}, while multi-vector late-interaction models such as ColBERT~\cite{khattab2020colbert} improve fine-grained token matching.

In agricultural RAG, KinyaColBERT~\cite{nzeyimana2025kinyacolbert} attributes low accuracy in Kinyarwanda to vocabulary coverage over architecture choice. Query-document vocabulary mismatch is an established retrieval challenge across domains, from general human-system communication~\cite{furnas1987vocabulary} and cross-genre retrieval~\cite{jurczyk2017crossgenre} to medical consumer health search~\cite{zeng2006exploring}, where lay queries routinely diverge from the formal vocabulary of the documents they need to retrieve.

In Bengali, BenLLM-Eval~\cite{kabir2024benllm} and TigerLLM~\cite{raihan2025tigerllm} evaluate LLM capabilities, while KrishokBondhu~\cite{ameen2026krishokbondhu} automates voice advisory and Farmer.Chat~\cite{singh2024farmerchat} extends AI-powered advisory to smallholder farmers across multiple languages. Closest to our setting, our own prior work~\cite{reza2026krishokchat} provides a QA generation dataset, and Hossain et al.~\cite{hossain2026cost} translate queries before retrieval. Unlike these generation-focused efforts, our work establishes a retrieval-only benchmark to isolate retriever architecture performance under controlled register and language shifts.

As summarized in Table~\ref{tab:positioning}, our benchmark is among the first to combine a low-resource language, architecture-isolated evaluation, retrieval-only measurement, controlled cross-lingual conditions, and provenance-grounded gold labels in a single design. For example, while MIRACL and Mr.\ TyDi provide diverse multilingual datasets, their standard evaluations do not isolate retrieval architectures across controlled intra-language register shifts (marked \xmark\ for Arch.\ isolated).

\begin{table}[t]
\caption{Positioning of our benchmark relative to closest prior benchmarks and systems.}
\label{tab:positioning}
\resizebox{\columnwidth}{!}{%
\begin{tabular}{@{}lccccc@{}}
\toprule
\textbf{Resource} & \textbf{Low-res.} & \textbf{Arch.} & \textbf{Retr.} & \textbf{Cross-} & \textbf{Prov.} \\
 & \textbf{lang.} & \textbf{isolated} & \textbf{only} & \textbf{lingual} & \textbf{ground.} \\
\midrule
BEIR~\cite{thakur2021beir}                  & \xmark & \cmark & \cmark & \xmark & \xmark \\
CRAG~\cite{yang2024crag}                    & \xmark & \xmark & \xmark & \xmark & \xmark \\
MIRACL~\cite{zhang2023miracl}               & \cmark & \xmark & \cmark & \cmark & \xmark \\
Mr.\ TyDi~\cite{zhang2021mr}                & \cmark & \xmark & \cmark & \cmark & \xmark \\
Amharic study~\cite{yeshambel2025dense}     & \cmark & \xmark & \cmark & \xmark & \xmark \\
Cross-lingual RAG~\cite{hossain2026cost}    & \cmark & \xmark & \xmark & \cmark & \xmark \\
Prior advisory benchmark~\cite{reza2026krishokchat} & \cmark & \xmark & \xmark & \xmark & \cmark \\
KinyaColBERT~\cite{nzeyimana2025kinyacolbert} & \cmark & \xmark & \cmark & \xmark & \xmark \\
\textbf{This work}                          & \cmark & \cmark & \cmark & \cmark & \cmark \\
\bottomrule
\end{tabular}%
}
\vspace{2pt}
\begin{minipage}{\columnwidth}
\footnotesize
Low-res.\ lang.\ = targets a low-resource language; Arch.\ isolated = evaluates multiple retrieval architectures in isolation from generation; Retr.\ only = a retrieval-only benchmark, with no downstream QA/generation layer required; Cross-lingual = includes a controlled cross-lingual query condition; Prov.\ ground.\ = gold labels are grounded in document-level provenance metadata.
\end{minipage}
\end{table}

\section{The Benchmark Collection}
\label{sec:benchmark}

\begin{figure}[t]
\centering
\includegraphics[width=\columnwidth, keepaspectratio]{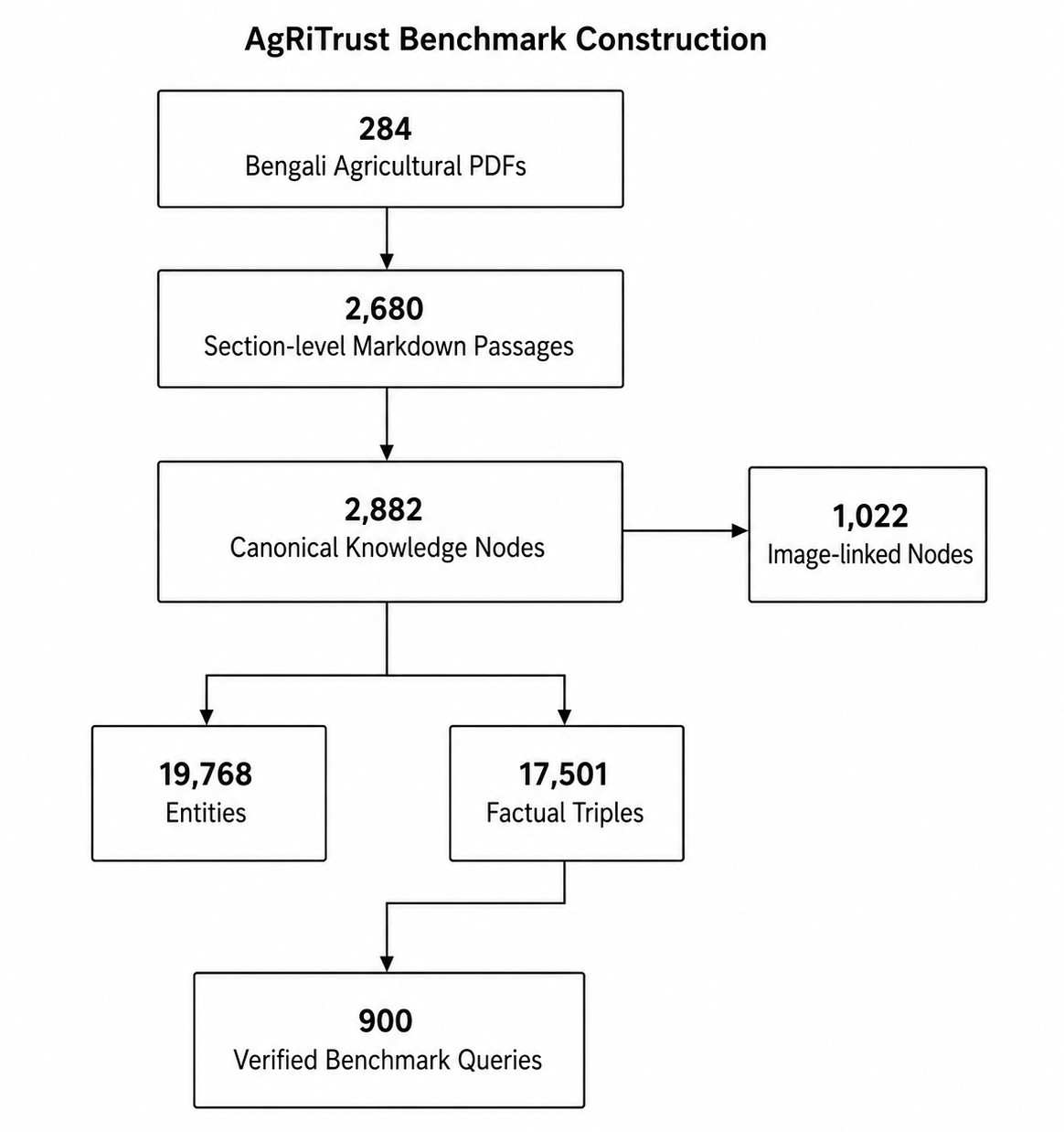}
\caption{Canonical knowledge node construction pipeline. Source documents are parsed, chunked by topic, and processed into structured, provenance-grounded JSON nodes.}
\Description{Flowchart illustrating the benchmark construction pipeline: 284 Bengali PDFs parsed to 2,680 section passages, yielding 2,882 canonical knowledge nodes (1,022 image-linked), 19,768 entities, 17,501 factual triples, and 900 verified benchmark queries.}
\label{fig:pipeline}
\end{figure}

We construct the benchmark around three requirements (Figure~\ref{fig:pipeline}): (i) document-level provenance, so every retrieval decision traces to a single source; (ii) knowledge-graph grounding, so queries are verified against structured facts instead of free text; and (iii) independent verification of node quality and query-document mappings, not gold labels accepted from a single automated pass.

\subsection{Source Materials}
We collect 284 Bengali agricultural PDFs from five government and research organizations (BRRI, IRRI, DAE, SRDI, MoA). The Ministry of Agriculture (MoA) and its extension arm (DAE) jointly contribute approximately 65\% of the corpus via national farming handbooks, while specialized institutes (BRRI, IRRI, SRDI) supply targeted diagnostic manuals. Many source documents contain scanned pages, complex layouts, tables, and embedded figures. We therefore convert each PDF into page-level Markdown using a layout-preserving OCR pipeline based on Mistral OCR, retaining document structure and page provenance for subsequent node construction. The resulting Markdown corpus (2,680 section-level passages) serves solely as an intermediate preprocessing artifact. While 73\% of this underlying raw text originates from our own prior QA benchmark~\cite{reza2026krishokchat}, this work's primary contribution is restructuring this text into a novel, provenance-grounded knowledge graph (100\% of KG entity and triple construction, node schema, and the full 900-query gold-mapping are new; only the underlying text and colloquial queries are inherited), establishing a dedicated retrieval benchmark where prior work left retrieval for future evaluation.

\textit{Corpus Diversity.} To ensure robust retrieval evaluation across the agricultural long tail, the extracted corpus spans 25 years of publications (1999 to 2024) encompassing five major document genres: national farming handbooks, diagnostic manuals, pesticide whitelists, recommendation cards, and extension guides. The resulting entity graph explicitly indexes 915 unique crops (spanning horticulture, agroforestry, and aquaculture), 704 disease variants, and 2,729 chemical or pesticide entities.

\subsection{Canonical Knowledge Node Construction}
\label{sec:node-construction}
Each knowledge node is a provenance-preserving retrieval unit spanning three layers: natural-language content, structured agricultural facts, and provenance (source, page, organization). Generation is bounded, not open-ended: extraction is constrained to information explicitly present in the source text, and provenance fields are injected deterministically, not model-generated. A representative node is given in Appendix~\ref{sec:appendix-b}; the full extraction and generation prompts are detailed in Appendix~\ref{sec:appendix-d}.

\textit{Deterministic Provenance Injection.} Citation metadata (publisher, source document, page range) is never LLM-generated: the generation prompt structurally forbids the model from producing citation fields, which are instead injected by a separate deterministic script directly from Markdown trace comments. This decouples provenance from semantic generation, making citation hallucination (fabricated page numbers or source names) structurally impossible rather than merely unlikely. A separate post-hoc verification pass cross-checks every extracted crop and entity mention against the source Markdown text, flagging any node whose entities do not appear in its cited passage for correction.

\textit{Chunking.} We segment text into 2,882 topic-coherent knowledge nodes, each corresponding to a single agricultural concept. Nodes average $\approx$1,180 characters, reflecting natural concept boundaries rather than fixed token windows, and are distributed across a 13-category agricultural taxonomy, ranging from Variety (695 nodes) and Cultivation Practice (570) down to a long tail such as Food Safety (18).

\textit{Entity and Triple Extraction.} We extract 19,768 entities (6.9/node) and 17,501 factual triples via schema-guided, bounded open information extraction. Per-node entity coverage exceeds 96\%.

\textit{Source-Grounded Automated Verification.} Node content (summary, symptoms, management, entities) was generated by Gemini-3.1-Flash-Lite from the source Markdown into a structured JSON schema. Each generated node was then independently verified by a separate model, GPT-5-Nano, which received both the node and its ground-truth source passage and scored the pair (0--5) along four criteria: faithfulness (every claim traceable to the source text), completeness, hallucination (unsupported content not present in source), and crop/entity relevance. Using a separate model for generation and verification reduces self-confirmation bias. Because the verifier sees the source passage, not just the node in isolation, its check is constrained to factual consistency and never becomes an open-ended quality rating. Across the resulting graph, entity coverage is 96.1\% (2,771 of 2,882 nodes contain at least one extracted entity) and factual triple coverage reaches 100\%.

\textit{Closed-Loop Refinement.} Nodes falling below threshold ($\leq$3 on any dimension; 283 of 2,882 nodes, 9.8\%) were returned to Gemini-3.1-Flash-Lite together with the verifier's specific error trace and regenerated against the same source passage, then re-verified. This repeated until each node passed threshold, yielding the final 2,882-node collection without discarding agricultural content.

\textit{Human Audit.} Because the automated stage already enforces source-level faithfulness, human review targeted a layer the automated check is not designed to catch: domain-appropriate terminology and information completeness under compression, judgments requiring agricultural expertise beyond simple text-source comparison. Three expert agricultural annotators independently evaluated a stratified random sample of 200 nodes on two dimensions: information completeness (critical dosage/withholding details dropped under compression) and technical terminology (formal Bengali agricultural terms vs.\ code-mixed transliterations). Annotators reached strong agreement (Fleiss' $\kappa{=}0.81$); 12 of 200 nodes (6.0\%) fell below consensus threshold and were corrected via direct manual edit against the source Markdown. Each node remains linked to its originating Markdown passage through deterministic provenance metadata, while image-containing nodes continue through the multimodal extension described in \S\ref{sec:image-nodes}.

\subsection{Image-Linked Nodes}
\label{sec:image-nodes}
Of the 2,882 canonical knowledge nodes, 1,022 (35.5\%) additionally carry a linked visual asset (a diagnostic photograph, variety chart, or table-image) captured via an \texttt{image\_refs} field on the same text node, under the identical provenance schema (\texttt{publisher}, \texttt{source\_document}, \texttt{source\_pages}) as the node's textual content. Each image reference additionally carries a figure-type label (one of 12 categories, e.g., \texttt{symptom\_close\_up}, \texttt{variety\_portrait}, \texttt{procedure\_illustration}) and a bilingual visual description (\texttt{visual\_description\_en/bn}) generated by a vision-language model, textualizing symptom color, shape, and texture rather than requiring a dedicated vision encoder at retrieval time. These fields extend the schema in Appendix~\ref{sec:appendix-c} but, like the image reference itself, are not separately indexed or evaluated in the present retrieval experiments (\S\ref{sec:results}).

This image-to-text linkage is motivated directly by the register gap identified in \S\ref{sec:register-gap}. Because farmer queries disproportionately describe visually observable symptoms (``leaves turning yellow'') instead of the formal entity names (\textit{Tungro virus}) used in source documents, image-linked nodes are the natural mechanism by which a future system could close exactly the severe retrieval gap this benchmark exposes (R@10=0.093 on farmer queries), rather than acting as a general-purpose multimodal extension. We report their structure in Appendix~\ref{sec:appendix-c} and leave dedicated visual retrieval to future work (\S\ref{sec:limitations}).

\subsection{Query Benchmark}
The benchmark contains 1,000 queries across three categories (Table~\ref{tab:stats}).

\textit{Farmer-anchored.} 400 farmer-anchored queries adapted from our own prior QA benchmark~\cite{reza2026krishokchat}. After three-annotator independent gold-node mapping, 300 achieved majority-vote consensus and are retained as answerable; the remaining 100 lacked majority-vote agreement among annotators (Fleiss' $\kappa$ for this subset: $<$0.4) and are excluded from retrieval evaluation. These 100 excluded queries constitute the \textit{low-agreement} subset; disagreement may reflect ambiguous queries or multiple valid gold documents rather than strict unanswerability.

\textit{KG-grounded (400).} 400 queries constructed directly from knowledge-graph triples (e.g., [crop, has\_disease, disease\_name]), targeting multi-hop relational retrieval. All 400 are verified by three annotators (Fleiss' $\kappa{=}0.78$) and included in the answerable set.

\textit{Safety (200).} All 200 queries test advisory failure modes (pesticide dosage, off-label chemical use, dangerous agronomic advice), each with a verified gold document. Safety queries use formal, precise language (chemical product codes, BRRI variety names) that closely mirrors document vocabulary: they show 1.6$\times$ higher lexical overlap with their gold document than farmer queries (\S\ref{sec:register-gap}), consistent with their high retrievability in \S\ref{sec:rq3}. All 200 are answerable and included in the 900-query evaluation set.

\textit{Independent Verification.} Farmer-anchored and safety queries were mapped by three annotators (Fleiss' $\kappa=0.72$, $n{=}600$). Majority vote determines the gold node; queries without majority agreement are excluded. KG-grounded gold nodes (400 queries) were assigned by a domain expert annotator and independently validated by three native Bengali-speaking annotators who evaluated each assignment on factual grounding, faithfulness, and relevance (Fleiss' $\kappa{=}0.78$, $n{=}400$). The final benchmark contains 900 answerable queries (300 farmer + 400 KG + 200 safety); the 100 excluded farmer queries are withheld as a low-agreement subset.

\begin{table}[t]
\caption{Benchmark dataset statistics.}
\label{tab:stats}
\resizebox{\columnwidth}{!}{%
\begin{tabular}{@{}lr@{}}
\toprule
\textbf{Property} & \textbf{Value} \\
\midrule
Source PDFs & 284 \\
Organizations & 5 (BRRI, IRRI, DAE, SRDI, MoA) \\
KG nodes & 2,882 \\
\quad Image-linked nodes & 1,022 (35.5\%) \\
Unique entities & 19,768 \\
Factual triples & 17,501 \\
\midrule
Queries (total generated) & 1,000 \\
\quad Farmer-anchored (answerable) & 300 \\
\quad Farmer-anchored (low-agreement) & 100 \\
\quad KG-grounded (answerable) & 400 \\
\quad Safety-critical (answerable) & 200 \\
\textbf{Answerable evaluation set} & \textbf{900} \\
\midrule
Inter-annotator $\kappa$ (farmer+safety) & 0.72 \\
Inter-annotator $\kappa$ (KG-grounded) & 0.78 \\
Entity coverage & $>$96\% \\
\bottomrule
\end{tabular}%
}
\end{table}

\subsection{Cross-Lingual Conditions}
\label{sec:crosslingual}
\label{sec:crossling}
To isolate language as an explanatory variable independent of retrieval architecture, we translate both the full query set and all 2,882 KG nodes into English, guided by a domain-specific agricultural glossary (312 domain-critical terms, cross-referenced against NCBI Taxonomy; 97\% back-translation equivalence on a 100-query spot check) to preserve technical terminology. This produces three controlled conditions on an otherwise identical benchmark: Bengali queries against the Bengali KB (BN$\to$BN), English queries against the Bengali KB (EN$\to$BN, cross-lingual), and English queries against the translated English KB (EN$\to$EN). Because the corpus content and evaluation procedure are held constant across conditions, the comparison isolates the effect of changing the language condition as closely as possible.

\section{Experimental Design}
\label{sec:experiment}

We evaluate five architectures and six embeddings not to advance retrieval methodology, but because architecture-isolated evaluation is itself part of the benchmark's contribution (\S\ref{sec:related}). Our evaluation is designed to isolate whether retrieval behavior is primarily determined by architecture, embedding model, or language condition. This approach reveals failure modes that remain invisible to single-architecture or aggregate-only evaluations. We organize primary experiments around four research questions: which architecture leads in Bengali (RQ1); whether this depends on the language boundary (RQ2); whether query type moderates performance (RQ3); and how embedding choice affects results (RQ4). We additionally conduct a configuration audit to verify that architecture comparisons are not artifacts of implementation settings. Analysis of observed failures is in \S\ref{sec:mechanisms}.

\subsection{Retrieval Architectures}
We evaluate five representative retrieval systems spanning four retrieval families: (i) \textbf{BM25}~\cite{robertson2009probabilistic} uses Okapi weighting ($k_1{=}1.5$, $b{=}0.75$) with character bigram plus full-word tokenization; (ii) \textbf{Dense (Gemini)} uses $L_2$-normalized 3,072-dim \texttt{gemini-embedding-001} vectors with exact inner-product search (FAISS IndexFlatIP) and asymmetric task types; (iii) \textbf{Dense (BGE-M3 Native)} uses 1,024-dim native dense embeddings as an open, retrieval-specific baseline; (iv) \textbf{ColBERT}~\cite{khattab2020colbert} applies BGE-M3 multi-vector scoring at 512-token passage granularity (to match the dense baselines' context window) with MaxSim; and (v) \textbf{Hybrid RRF} fuses BM25 and Gemini dense rankings via Reciprocal Rank Fusion ($k{=}60$, top-100 candidates each). All architectures retrieve from the identical 2,882-node benchmark corpus introduced in \S\ref{sec:benchmark}; only the retrieval mechanism differs. Full configuration settings are in Appendix~\ref{sec:appendix-a}.

\subsection{Embedding Robustness}
To separate architectural effects from embedding quality, we evaluate six models spanning 384--4,096 dimensions (Table~\ref{tab:embeddings}) on the same Bengali query set under identical retrieval code, isolating embedding choice as the only varying factor.

\subsection{Metrics}
\textit{Retrieval (L1).} We report R@1, R@5, R@10 (hereafter R@k), MRR, and nDCG@10 against gold source nodes for the 900 answerable queries. A retrieved node counts as a hit only if its exact node identifier matches the provenance-verified gold node (serving as the strict retrieval document): a deterministic criterion admitting no partial credit. For context, the random baseline R@10 for a corpus of this size is $10/2{,}882 \approx 0.003$.

\textit{Separation score.} Beyond Recall@$k$, we measure how well embeddings separate the gold node from the remainder of the corpus. For query $q_i$ with gold node $g_i$, let $s_{\text{gold}} = \cos(E(q_i), E(g_i))$ and $s_{\text{neg}}$ be the mean cosine similarity between $E(q_i)$ and all 2,881 non-gold nodes. The separation score is the mean difference ($s_{\text{gold}} - s_{\text{neg}}$) across all $N$ evaluated queries. A score of zero indicates random ranking behaviour; negative values indicate the gold node ranks below the corpus average.

\subsection{Statistical Validation}
We compare retrieval architectures using BCa bootstrap confidence intervals (10,000 resamples). Pairwise significance is assessed via paired Wilcoxon signed-rank tests with Holm-Bonferroni correction across 10 architecture pairs in Table~\ref{tab:l1results}; per-category and per-embedding tables are reported descriptively. Significance levels ($p<0.001$ for BM25 vs.\ Dense; $p<0.01$ for Hybrid RRF) and 95\% CIs are reported inline in \S\ref{sec:rq1}.

\section{Results}
\label{sec:results}

Each subsection answers one research question using the 900 answerable Bengali queries unless otherwise noted. All reported numbers are from the post-audit verification run; Section~\ref{sec:mechanisms} discusses the observed failure patterns.

\subsection{RQ1: Benchmark Difficulty and Architecture Comparison}
\label{sec:rq1}
Table~\ref{tab:l1results} presents R@10 across 900 answerable Bengali queries (random baseline R@10$\approx$0.003). Hybrid RRF achieves the highest overall R@10 (0.539): lexical and dense signals provide complementary coverage. Among single-method retrievers, BM25 leads (R@10=0.506, 95\% CI: [0.474, 0.538]), outperforming dense Gemini-001 (0.464, 95\% CI: [0.432, 0.497]) by 9\% ($p<0.001$, Wilcoxon signed-rank test). Late-interaction ColBERT (BGE-M3, 512 tokens) reaches R@10=0.487, outperforming both single-vector dense variants (0.464, 0.408). Broken out by query category (Table~\ref{tab:bycategory}), ColBERT is the best single architecture on KG-grounded queries (R@10=0.600, vs.\ Dense 0.489, BM25 0.478) and reduces the gap on farmer queries relative to Dense (0.210 vs.\ 0.093).

\begin{table}[t]
\caption{L1 Retrieval on 900 Answerable Bengali Queries (BN$\to$BN). Bold = best overall; underline = best single-method.}
\label{tab:l1results}
\resizebox{\columnwidth}{!}{%
\begin{tabular}{@{}lccccc@{}}
\toprule
\textbf{Architecture} & \textbf{R@1} & \textbf{R@5} & \textbf{R@10} & \textbf{MRR} & \textbf{nDCG@10} \\
\midrule
Hybrid RRF (Gemini+BM25) & .291 & \textbf{.466} & \textbf{.539} & \textbf{.551} & \textbf{.461} \\
BM25 (Sparse)            & .205 & .405 & \underline{.506} & .481 & .407 \\
Dense (Gemini-001)       & \underline{\textbf{.320}} & \underline{.431} & .464 & \underline{.514} & \underline{.431} \\
Dense (BGE-M3 Native)    & .281 & .369 & .408 & .432 & .370 \\
ColBERT (BGE-M3)$^\ddagger$ & .255 & .414 & .487 & .324 & .416 \\
\bottomrule
\end{tabular}%
}
\vspace{2pt}
\begin{minipage}{\columnwidth}
\footnotesize
$\ddagger$ ColBERT evaluated at \texttt{max\_seq\_length=512}, matching the dense baselines' context window.
\end{minipage}
\end{table}

\begin{table}[t]
\caption{Cross-Lingual Comparison (95\% BCa CIs for R@10 in brackets; 900 queries).}
\label{tab:crosslingual}
\resizebox{\columnwidth}{!}{%
\begin{tabular}{@{}lccc@{}}
\toprule
\textbf{Setting} & \textbf{Dense (Gemini)} & \textbf{BM25} & \textbf{Winner} \\
\midrule
Bengali (BN$\to$BN)       & .464 {\scriptsize[.43, .50]} & \textbf{.506} {\scriptsize[.47, .54]} & BM25 (+9\%) \\
Cross-lingual (EN$\to$BN) & \textbf{.425} {\scriptsize[.39, .46]} & .004 {\scriptsize[.00, .01]} & Dense ($\approx$100$\times$) \\
English (EN$\to$EN)       & \textbf{.442} {\scriptsize[.41, .47]} & .384 {\scriptsize[.35, .42]} & Dense (+15\%) \\
\bottomrule
\end{tabular}%
}
\end{table}

\subsection{RQ2: Is architectural failure determined by language mismatch?}
Table~\ref{tab:crosslingual} examines whether language boundary crossing changes retrieval behaviour independent of architecture. Within the same language, BM25 leads in Bengali (0.506 vs.\ 0.464) while Dense leads in English (0.442 vs.\ 0.384): neither architecture is universally dominant. The results diverge markedly in the cross-lingual setting (English queries $\to$ Bengali corpus), where BM25's exact lexical matching declines sharply across script boundaries (R@10: 0.506 $\to$ 0.004, a 99\% drop) and becomes effectively unusable. Multilingual dense embeddings, by contrast, largely preserve the script boundary (R@10: 0.464 $\to$ 0.425, only an 8\% drop). Architecture choice cannot be separated from the language scenario it is deployed in.

\subsection{RQ3: Does the query type moderate retrieval performance?}
\label{sec:rq3}
Dense retrieval's aggregate Bengali score masks large differences across query types (Table~\ref{tab:bycategory}). Dense retrieval (Gemini) performs poorly on colloquial farmer queries (R@10=0.093) but reaches high R@10 on formal safety queries (0.970), a difference driven entirely by query register. Safety queries function as near-direct entity lookups, with little inferential gap to close between query and gold document (\S\ref{sec:register-gap}); farmer queries instead require bridging the symptom-to-entity register gap (\S\ref{sec:register-gap}). BM25 is comparatively stable across categories: farmer (0.523), safety (0.539), KG (0.478). On KG-grounded queries dense leads BM25 by a narrow margin (0.489 vs.\ 0.478); BM25's same-language advantage over dense comes from consistent performance across colloquial and formal query types, in contrast to the large gap between the two query types for dense retrieval.

\begin{table}[t]
\caption{R@10 by query category (95\% BCa CIs in brackets; 900 verified queries; Farmer=300, Safety=200, KG-grounded=400).}
\label{tab:bycategory}
\resizebox{\columnwidth}{!}{%
\begin{tabular}{@{}lcccc@{}}
\toprule
\textbf{Architecture} & \textbf{Farmer} & \textbf{Safety} & \textbf{KG-Ground.} & \textbf{Farm.+Safe.} \\
 & \textit{(colloquial)} & \textit{(formal)} & \textit{(formal)} & \textit{(combined)} \\
\midrule
Dense (Gemini) & .093 {\scriptsize[.06,.13]} & \textbf{.970} {\scriptsize[.94,.99]} & .489 {\scriptsize[.44,.54]} & .444 {\scriptsize[.40,.49]} \\
BM25 & \textbf{.523} {\scriptsize[.47,.58]} & .539 {\scriptsize[.47,.60]} & .478 {\scriptsize[.43,.53]} & \textbf{.529} {\scriptsize[.48,.57]} \\
ColBERT (BGE-M3)$^\ddagger$ & .210 {\scriptsize[.17,.26]} & .675 {\scriptsize[.60,.74]} & \textbf{.600} {\scriptsize[.55,.65]} & .396 {\scriptsize[.35,.44]} \\
\bottomrule
\end{tabular}%
}
\end{table}

\begin{figure}[t]
\centering
\begin{tikzpicture}
\begin{axis}[
    ybar,
    bar width=6.5pt,
    width=\columnwidth,
    height=4.3cm,
    ymin=0, ymax=1.08,
    ylabel={R@10},
    ylabel style={font=\footnotesize},
    symbolic x coords={Farmer, Safety, KG-grounded},
    xtick=data,
    tick label style={font=\footnotesize},
    enlarge x limits=0.28,
    legend style={at={(0.5,1.22)}, anchor=north, legend columns=-1, font=\scriptsize, draw=none},
    ytick={0,0.2,0.4,0.6,0.8,1.0},
    axis line style={gray!60},
    ymajorgrids, grid style={gray!25},
]
\addplot[fill=orange!70!black, draw=none] coordinates {(Farmer,0.093) (Safety,0.970) (KG-grounded,0.489)};
\addplot[fill=blue!55!black, draw=none]   coordinates {(Farmer,0.523) (Safety,0.539) (KG-grounded,0.478)};
\addplot[fill=teal!70!black, draw=none]   coordinates {(Farmer,0.210) (Safety,0.675) (KG-grounded,0.600)};
\legend{Dense (Gemini), BM25, ColBERT}
\end{axis}
\end{tikzpicture}
\caption{R@10 by query register, plotted from Table~\ref{tab:bycategory}. Dense retrieval is sharply bimodal --- it nearly fails on colloquial farmer queries and saturates on formal safety queries --- while BM25 stays comparatively stable across registers.}
\Description{Bar chart showing R@10 across Farmer, Safety, and KG-grounded registers for Dense Gemini, BM25, and ColBERT.}
\label{fig:registergap}
\end{figure}
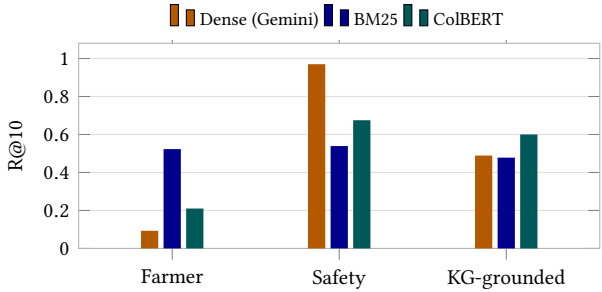

\subsection{RQ4: How do different embedding models compare?}
Evaluating six embedding models spanning 384--4,096 dimensions (Table~\ref{tab:embeddings}) shows that retrieval-specific training dominates model scale: Gemini-001 (0.464) and BGE-M3 (0.408) outperform larger models like Qwen3-8B (0.241). Notably, BGE-M3 yields the highest mean cosine separation (+0.133 vs.\ Gemini's +0.127) despite ranking second on Recall@10 (.408 vs.\ .464); mean separation rewards models with confident average alignment, whereas top-$k$ recall penalizes a long tail of complete retrieval misses. General-purpose sentence-similarity encoders (MPNet, MiniLM) fail (R@10 $<$ 0.01), yielding negative separation scores that indicate gold nodes are ranked below corpus average.

\begin{table}[t]
\caption{Multi-Embedding Analysis (95\% BCa CIs for R@10 in brackets; 900 queries, BN$\to$BN). Random baseline: separation = 0.}
\label{tab:embeddings}
\small
\begin{tabular}{@{}lrrc@{}}
\toprule
\textbf{Model} & \textbf{Dim} & \textbf{Separation} & \textbf{R@10} \\
\midrule
Gemini-embedding-001    & 3,072 & $+0.127$ & \textbf{.464} {\scriptsize[.43,.50]} \\
BGE-M3 (native proj.)   & 1,024 & $\mathbf{+0.133}$ & .408 {\scriptsize[.38,.44]} \\
E5-Large (multilingual) & 1,024 & $+0.037$ & .284 {\scriptsize[.26,.32]} \\
Qwen3 Embedding 8B      & 4,096 & $+0.108$ & .241 {\scriptsize[.21,.27]} \\
MPNet (paraphrase)      &   768 & $-0.010$ & .009 {\scriptsize[.00,.02]} \\
MiniLM (paraphrase)     &   384 & $-0.039$ & .005 {\scriptsize[.00,.01]} \\
\bottomrule
\end{tabular}
\end{table}

\textit{Configuration Audit.} To verify that architecture comparisons are not implementation artifacts, we audited configuration sensitivity across embedding APIs and passage granularity. Using a sentence-similarity task type instead of the asymmetric \texttt{RETRIEVAL\_QUERY}/\texttt{RETRIEVAL\_DOCUMENT} encoding reduced Gemini Dense R@10 from 0.464 to 0.063, a 7$\times$ drop. Passage granularity is a second, independent configuration axis: at the field-standard default of 128 tokens, Bengali knowledge nodes (mean $\approx$1,180 characters) are truncated by 95--99\%, and ColBERT falls below even the weaker dense baseline (R@10=0.376 vs.\ BGE-M3's 0.408); increasing the context window to 512 tokens lifts ColBERT to R@10=0.487 (Table~\ref{tab:l1results}), reversing this ranking. Two independent default settings each changed reported architecture rankings by a wide margin. Configuration auditing should therefore precede architecture comparison in low-resource retrieval settings. Full configuration details are in Appendix~\ref{sec:appendix-a}.

\section{Analysis of Retrieval Failures}
\label{sec:mechanisms}

This section explains the performance patterns from Section~\ref{sec:results} through two lenses: the query-document register gap (\S\ref{sec:register}), and the complementary failure patterns between BM25 and dense retrieval (\S\ref{sec:complementarity}).

\subsection{Register Gap Analysis}
\label{sec:register}
\label{sec:register-gap}
\label{sec:lexgap}
Token-level Jaccard similarity across all 900 query-gold pairs reveals a near-universal lexical gap: \textbf{96.4\%} of queries have Jaccard $<$~0.10 with their gold document (mean: 0.044; maximum: 0.172). This confirms that benchmark evaluation requires semantic matching, not direct surface overlap.

However, surface overlap varies substantially across query registers: safety queries achieve a mean Jaccard of 0.055 vs.\ 0.034 for farmer queries ($\mathbf{1.6{\times}}$ higher alignment). Verbatim entity inspection explains dense retrieval's collapse on colloquial farmer queries (R@10${=}0.093$): only 3.5\% of gold document entity names appear verbatim in farmer queries, while 92\% are entirely absent. Farmers describe observable symptoms (\emph{``leaves turning yellow''}), whereas authoritative documents encode formal scientific entities (\emph{Tungro virus}) (illustrated in Table~\ref{tab:example-query}). Safety queries use precise chemical codes and variety names that closely match document vocabulary, so dense retrievers achieve high accuracy (0.970) on this category.

\begin{table}[t]
\centering
\small
\caption{Illustrative register gap: a farmer-anchored query and its gold node (English glosses of Bengali text in brackets).}
\label{tab:example-query}
\begin{tabular}{@{}p{0.95\linewidth}@{}}
\toprule
\textbf{Farmer query (colloquial):} [My chickpea plants' leaves are turning yellow and drooping, what should I do?] \\[2pt]
\textbf{Gold node title:} [Chickpea crop disease control and remedies] (\texttt{B7\_CH2\_CHHOLA\_0006\_N1}) \\[2pt]
\textbf{Gold node entity:} \emph{wilt disease} (formal pathological term; not present in query) \\[2pt]
\textbf{Token-level Jaccard:} 0.03 (below 0.10 near-universal gap threshold, \S\ref{sec:lexgap}) \\
\bottomrule
\end{tabular}
\end{table}

\subsection{Failure Complementarity and Hybrid Attribution}
\label{sec:complementarity}
Hybrid RRF's gain over either single-method retriever is explained by complementary failure patterns between BM25 and Dense retrieval at the category level (Table~\ref{tab:bycategory}). BM25 is comparatively stable across query registers (farmer 0.523, safety 0.539, KG-grounded 0.478), while Dense is sharply bimodal (farmer 0.093, safety 0.970). Because the two architectures reach their strongest and weakest performance on largely different query registers, fusing their rankings recovers queries that either method alone would miss, consistent with Hybrid RRF's overall gain over both single methods (R@10${=}0.539$ vs.\ 0.506 and 0.464). Persistent shared difficulty concentrates in colloquial farmer queries and multi-hop KG-grounded queries, where neither lexical nor semantic matching reliably surfaces the gold document, consistent with the symptom-to-entity register mismatch identified above as the primary barrier in low-resource retrieval.

\section{Discussion}

\subsection{Methodological Implications}
Evaluating low-resource retrieval with aggregate scores alone hides important failure modes. Dense retrieval R@10 drops from 0.970 on formal safety queries to 0.093 on colloquial farmer queries, corroborating concurrent findings on stratified retrieval evaluation~\cite{klearman2026coverage}. Across language conditions, BM25 leads within Bengali while dense retrieval leads under cross-lingual conditions. Our configuration audit additionally shows that two independent default settings, an embedding-API task type and passage-length truncation, each changed reported architecture rankings by a wide margin; configuration auditing should therefore precede architectural evaluation~\cite{wang2024multilinguale5}.

\subsection{Limitations and Ethics}
\textbf{Limitations.} \label{sec:limitations}
Evaluation is limited to agricultural advisory and to first-stage zero-shot retrievers; learned sparse models (e.g., SPLADE) and cross-encoder rerankers are left to future work. Machine-translated English queries may carry subtle lexical shifts despite 97\% back-translation equivalence on a 100-query random sample (\S\ref{sec:crossling}); future work should expand translation validation to the full corpus to systematically characterize terminology preservation. However, BM25's near-total collapse under cross-lingual querying (R@10${=}0.004$) reflects an expected consequence of exact lexical matching across non-cognate scripts, since the cross-lingual condition compares English queries directly against the untranslated Bengali corpus; this result is unlikely to be a translation artifact. Image-linked nodes (Appendix~\ref{sec:image-node-appendix}) attach a visual reference to a subset of text nodes but are not separately embedded or evaluated as a distinct retrieval modality in this work. A human audit of 200 nodes found 12 requiring correction, so a small fraction of unresolved failure queries may reflect annotation artifacts in the benchmark itself. Source documents span 1999--2024. The benchmark evaluates whether a retriever surfaces the correct authoritative source document, not whether that document's recommendations remain current; deploying any system built on this corpus for live advisory would require a separate regulatory-currency check independent of retrieval accuracy. Finally, while reliance on proprietary APIs for initial query generation poses a reproducibility risk as models deprecate, the resulting benchmark is released as a static, version-controlled JSON corpus, supporting stable retrieval evaluations over time.

\textbf{Ethics and broader impact.} Annotations were completed by three compensated domain-expert annotators with informed consent. The benchmark evaluates retrieval performance to motivate model improvement in low-resource domain advisory and contains no personally identifiable information. Given the safety-critical query subset, retrieval accuracy does not certify advisory correctness for unsupervised deployment; systems evaluated on this benchmark should retain human oversight for high-stakes recommendations.

\section{Conclusion}
We presented a provenance-grounded benchmark for Bengali agricultural retrieval built from authoritative government publications through canonical knowledge nodes, image-linked resources, and register- and language-stratified evaluation. Beyond providing a reproducible benchmark, our experiments show that retrieval behaviour depends jointly on linguistic register, language boundary, and implementation configuration rather than retrieval architecture alone. These findings provide a stronger empirical basis for evaluating future low-resource retrieval and RAG systems.

\appendix

\section{Reproducibility Checklist}
\label{app:reproducibility}
\label{sec:appendix-a}
\begin{itemize}[leftmargin=*, itemsep=0pt, topsep=1pt]
\raggedright
\item \textbf{Data:} 284 source PDFs, 2,882 knowledge nodes, 1,000 queries (900 with verified gold-node mappings; 100 low-agreement queries released for future work). Available at \url{https://huggingface.co/datasets/RaiyanKhaan/AgriTrust-RAG}.
\item \textbf{BM25:} Okapi ($k_1{=}1.5$, $b{=}0.75$), char-bigram + full-word tokenization.
\item \textbf{Dense (Gemini):} \texttt{gemini-embedding-001}, 3,072-dim, asymmetric \texttt{RETRIEVAL\_QUERY}/\texttt{RETRIEVAL\_DOCUMENT} task types.
\item \textbf{Dense (BGE-M3):} Native projection head, 1,024-dim, $L_2$-normalized.
\item \textbf{ColBERT:} BGE-M3 multi-vector, 512-token sequence length, MaxSim scoring.
\item \textbf{Hybrid RRF:} Reciprocal Rank Fusion ($k{=}60$, top-100 candidates per method).
\item \textbf{Audit Variants:} sentence-similarity task type; 128-token passage truncation.
\item \textbf{Stats:} BCa bootstrap (10,000 resamples), paired Wilcoxon signed-rank tests with Holm-Bonferroni correction.
\end{itemize}

\section{Node Representation}
\label{app:node-example}
\label{app:extraction-protocol}
\label{sec:appendix-b}
A representative canonical knowledge node (trimmed; full schema released with benchmark code), showing the three-layer structure (natural-language content, structured facts, deterministic provenance):

\begin{tcolorbox}[colback=gray!5, colframe=gray!50, boxrule=0.3pt, arc=1pt,
  left=3pt, right=3pt, top=2pt, bottom=2pt, fontupper=\scriptsize\ttfamily]
\{ "category": "disease",\\
\ "title\_bn": "[Chickpea crop disease control and remedies]",\\
\ "summary": "[Guidance on wilt disease control via seed treatment,\\
\ \ variety selection, and infected plant management]",\\
\ "key\_points": [\\
\ \ "[Treat seeds with Provex-200 WP at 2.5-3.0 g/kg rate]",\\
\ \ "[Grow resistant varieties: BARI Chickpea-5, -9, -10]"\\
\ ],\\
\ "structured": \{"symptoms": ["wilt disease"],\\
\ \ "management": ["seed treatment (Provex-200 WP)", "fungicide spray"],\\
\ \ "prevention": ["use resistant varieties"]\},\\
\ "entities": \{"chemicals": ["Provex-200 WP"],\\
\ \ "diseases": ["wilt disease"], "crops": ["chickpea"]\},\\
\ "generation\_model": "gemini-3.1-flash-lite",\\
\ "\_provenance\_layer": \{"node\_id": "B7\_CH2\_CHHOLA\_0006\_N1",\\
\ \ "publisher": "BARI",\\
\ \ "source\_document": "krishiProjuktiHatboi\_10.pdf",\\
\ \ "source\_pages": [112, 112]\} \}
\end{tcolorbox}
{\footnotesize Field values in brackets are English translations of the stored UTF-8 Bengali text.}

\section{Image-Linked Node Structure}
\label{sec:image-node-appendix}
\label{sec:appendix-c}
Nodes with a linked visual asset extend the standard schema
(Appendix~\ref{app:node-example}) with an \texttt{image\_refs} field,
without introducing a separate node type:

\begin{tcolorbox}[colback=gray!5, colframe=gray!50, boxrule=0.3pt, arc=1pt,
  left=3pt, right=3pt, top=2pt, bottom=2pt, fontupper=\scriptsize\ttfamily]
\{ ... \\
\ "image\_refs": [\\
\ \ \{ "caption\_en": "[Chickpea leaf showing early wilt symptoms]",\\
\ \ \ "image\_type": "diagnostic\_photo" \}\\
\ ], ... \}
\end{tcolorbox}
{\footnotesize All other fields (\texttt{node\_id}, \texttt{category}, \texttt{title\_bn}, \texttt{\_provenance\_layer}) are unchanged from the node shown in Appendix~\ref{app:node-example}.}

In total, 1,022 of 2,882 nodes (35.5\%) carry one or more image references. Image references have not undergone the three-annotator human audit (\S\ref{sec:node-construction}) applied to node text, and we report them as a structural resource, not an independently verified evaluation asset (\S\ref{sec:limitations}).

\section{Node Extraction Protocol}
\label{app:prompt}
\label{sec:appendix-d}
The canonical nodes in the benchmark are constructed via a two-stage generative pipeline bounded by deterministic checks. 

First, an automated extraction stage identifies critical entities (crops, diseases, agricultural chemicals) from the source Markdown passage and stores them in a temporary registry. The extraction prompt enforces exact surface-form matching and forbids hallucination:

\begin{tcolorbox}[colback=gray!5, colframe=gray!50, boxrule=0.3pt, arc=1pt, left=2pt, right=2pt, top=1pt, bottom=1pt, fontupper=\scriptsize\ttfamily]
\#\#\# Strict Bounding: Extract ONLY entities explicitly mentioned in the text. Do not infer or add external knowledge. If a category is missing, return an empty list.\\
\#\#\# Surface Form: Extract the verbatim string exactly as it appears in the Bengali text.\\
\#\#\# Categories: "crops", "diseases\_and\_pests", "chemicals".\\
\#\#\# Output: Emit a valid JSON object ONLY.
\end{tcolorbox}

Second, the Markdown passage is converted into structured JSON node(s) by Gemini-3.1-Flash-Lite. Following generation, a deterministic script verifies that the entities extracted in Stage 1 exactly match those present in the generated node. Provenance metadata is then injected purely deterministically. The abridged generation instruction enforces 1-to-N dynamic chunking and semantic bounding:

\begin{tcolorbox}[colback=gray!5, colframe=gray!50, boxrule=0.3pt, arc=1pt, left=2pt, right=2pt, top=1pt, bottom=1pt, fontupper=\scriptsize\ttfamily]
\#\#\# Strict Domain Bounding: The generated node(s) must only contain information from the specific .md document provided... Never claim a crop or topic that the trace and body do not support.\\
\#\#\# Faithfulness: Every factual claim must come from the section you were given. You do not add facts, doses, chemical names, or yields from memory. If you cannot ground a field in the text, leave it blank.\\
\#\#\# How many nodes: One coherent topic $\rightarrow$ one node. Several distinct sub-topics (separate varieties, a disease and its treatment) $\rightarrow$ one node each. Each node must be a distinct, non-redundant unit.\\
\#\#\# Category: Pick the single best from: variety, disease, pest, fertilizer, cultivation\_practice, post\_harvest, seed\_tech, machinery, irrigation, ipm, cropping\_system, food\_safety, general.\\
\#\#\# Schema Enforcement: Output a JSON array conforming to the node schema with fields "title\_bn", "summary", "key\_points", "structured", "entities".
\end{tcolorbox}

\bibliographystyle{ACM-Reference-Format}
\bibliography{references}


\begin{thebibliography}{27}


\ifx \showCODEN    \undefined \def \showCODEN     #1{\unskip}     \fi
\ifx \showISBNx    \undefined \def \showISBNx     #1{\unskip}     \fi
\ifx \showISBNxiii \undefined \def \showISBNxiii  #1{\unskip}     \fi
\ifx \showISSN     \undefined \def \showISSN      #1{\unskip}     \fi
\ifx \showLCCN     \undefined \def \showLCCN      #1{\unskip}     \fi
\ifx \shownote     \undefined \def \shownote      #1{#1}          \fi
\ifx \showarticletitle \undefined \def \showarticletitle #1{#1}   \fi
\ifx \showURL      \undefined \def \showURL       {\relax}        \fi
\providecommand\bibfield[2]{#2}
\providecommand\bibinfo[2]{#2}
\providecommand\natexlab[1]{#1}
\providecommand\showeprint[2][]{arXiv:#2}

\bibitem[Ameen et~al\mbox{.}(2026)]%
        {ameen2026krishokbondhu}
\bibfield{author}{\bibinfo{person}{Mohd~Ruhul Ameen}, \bibinfo{person}{Akif
  Islam}, \bibinfo{person}{Farjana Aktar}, {and} \bibinfo{person}{M~Saifuzzaman
  Rafat}.} \bibinfo{year}{2026}\natexlab{}.
\newblock \showarticletitle{KrishokBondhu: A Retrieval-Augmented Voice-Based
  Agricultural Advisory Call Center for Bengali Farmers}. In
  \bibinfo{booktitle}{\emph{2026 IEEE 2nd International Conference on Quantum
  Photonics, Artificial Intelligence \& Networking (QPAIN)}}. IEEE,
  \bibinfo{pages}{1--6}.
\newblock


\bibitem[Eberhard et~al\mbox{.}(2024)]%
        {eberhard2024ethnologue}
\bibfield{author}{\bibinfo{person}{David~M. Eberhard}, \bibinfo{person}{Gary~F.
  Simons}, {and} \bibinfo{person}{Charles~D. Fennig}.}
  \bibinfo{year}{2024}\natexlab{}.
\newblock \bibinfo{title}{Ethnologue: Languages of the World}.
\newblock
  \bibinfo{howpublished}{\url{https://www.ethnologue.com/language/ben/}}.
\newblock
\shownote{Online version}.
\newblock


\bibitem[Friel et~al\mbox{.}(2024)]%
        {friel2024ragbench}
\bibfield{author}{\bibinfo{person}{Robert Friel}, \bibinfo{person}{Masha
  Belyi}, {and} \bibinfo{person}{Atindriyo Sanyal}.}
  \bibinfo{year}{2024}\natexlab{}.
\newblock \showarticletitle{Ragbench: Explainable benchmark for
  retrieval-augmented generation systems}.
\newblock \bibinfo{journal}{\emph{arXiv preprint arXiv:2407.11005}}
  (\bibinfo{year}{2024}).
\newblock


\bibitem[Furnas et~al\mbox{.}(1987)]%
        {furnas1987vocabulary}
\bibfield{author}{\bibinfo{person}{George~W. Furnas},
  \bibinfo{person}{Thomas~K. Landauer}, \bibinfo{person}{Louis~M. Gomez}, {and}
  \bibinfo{person}{Susan~T. Dumais}.} \bibinfo{year}{1987}\natexlab{}.
\newblock \showarticletitle{The vocabulary problem in human-system
  communication}.
\newblock \bibinfo{journal}{\emph{Commun. ACM}} \bibinfo{volume}{30},
  \bibinfo{number}{11} (\bibinfo{year}{1987}), \bibinfo{pages}{964--971}.
\newblock


\bibitem[Hong et~al\mbox{.}(2026)]%
        {hong2026improving}
\bibfield{author}{\bibinfo{person}{Seongtae Hong}, \bibinfo{person}{Youngjoon
  Jang}, \bibinfo{person}{Jungseob Lee}, \bibinfo{person}{Hyeonseok Moon},
  {and} \bibinfo{person}{Heuiseok Lim}.} \bibinfo{year}{2026}\natexlab{}.
\newblock \showarticletitle{Improving semantic proximity in information
  retrieval through cross-lingual alignment}. In \bibinfo{booktitle}{\emph{The
  Fourteenth International Conference on Learning Representations}}.
\newblock


\bibitem[Hossain et~al\mbox{.}(2026)]%
        {hossain2026cost}
\bibfield{author}{\bibinfo{person}{Md~Asif Hossain}, \bibinfo{person}{Nabil
  Subhan}, \bibinfo{person}{Mantasha~Rahman Mahi}, {and}
  \bibinfo{person}{Jannatul~Ferdous Nabila}.} \bibinfo{year}{2026}\natexlab{}.
\newblock \showarticletitle{Cost-Efficient Cross-Lingual Retrieval-Augmented
  Generation for Low-Resource Languages: A Case Study in Bengali Agricultural
  Advisory}.
\newblock \bibinfo{journal}{\emph{arXiv preprint arXiv:2601.02065}}
  (\bibinfo{year}{2026}).
\newblock


\bibitem[Jurczyk and Choi(2017)]%
        {jurczyk2017crossgenre}
\bibfield{author}{\bibinfo{person}{Tomasz Jurczyk} {and}
  \bibinfo{person}{Jinho~D. Choi}.} \bibinfo{year}{2017}\natexlab{}.
\newblock \showarticletitle{Cross-genre Document Retrieval: Matching between
  Conversational and Formal Writings}. In \bibinfo{booktitle}{\emph{Proceedings
  of the First Workshop on Building Linguistically Generalizable NLP Systems}}.
  \bibinfo{publisher}{Association for Computational Linguistics}.
\newblock


\bibitem[Kabir et~al\mbox{.}(2024)]%
        {kabir2024benllm}
\bibfield{author}{\bibinfo{person}{Mohsinul Kabir},
  \bibinfo{person}{Mohammed~Saidul Islam}, \bibinfo{person}{Md~Tahmid~Rahman
  Laskar}, \bibinfo{person}{Mir~Tafseer Nayeem}, \bibinfo{person}{M~Saiful
  Bari}, {and} \bibinfo{person}{Enamul Hoque}.}
  \bibinfo{year}{2024}\natexlab{}.
\newblock \showarticletitle{BenLLM-eval: A comprehensive evaluation into the
  potentials and pitfalls of large language models on Bengali NLP}. In
  \bibinfo{booktitle}{\emph{Proceedings of the 2024 Joint International
  Conference on Computational Linguistics, Language Resources and Evaluation
  (LREC-COLING 2024)}}. \bibinfo{pages}{2238--2252}.
\newblock


\bibitem[Khattab and Zaharia(2020)]%
        {khattab2020colbert}
\bibfield{author}{\bibinfo{person}{Omar Khattab} {and} \bibinfo{person}{Matei
  Zaharia}.} \bibinfo{year}{2020}\natexlab{}.
\newblock \showarticletitle{Colbert: Efficient and effective passage search via
  contextualized late interaction over bert}. In
  \bibinfo{booktitle}{\emph{Proceedings of the 43rd International ACM SIGIR
  conference on research and development in Information Retrieval}}.
  \bibinfo{pages}{39--48}.
\newblock


\bibitem[Klearman et~al\mbox{.}(2026)]%
        {klearman2026coverage}
\bibfield{author}{\bibinfo{person}{Andrew Klearman}, \bibinfo{person}{Radu
  Revutchi}, \bibinfo{person}{Rohin Garg}, \bibinfo{person}{Rishav
  Chakravarti}, \bibinfo{person}{Samuel~Marc Denton}, {and}
  \bibinfo{person}{Yuan Xue}.} \bibinfo{year}{2026}\natexlab{}.
\newblock \showarticletitle{Coverage, Not Averages: Semantic Stratification for
  Trustworthy Retrieval Evaluation}.
\newblock \bibinfo{journal}{\emph{arXiv preprint arXiv:2604.20763}}
  (\bibinfo{year}{2026}).
\newblock


\bibitem[Lee et~al\mbox{.}(2026)]%
        {lee2026clear}
\bibfield{author}{\bibinfo{person}{Seungyoon Lee}, \bibinfo{person}{Minhyuk
  Kim}, \bibinfo{person}{Seongtae Hong}, \bibinfo{person}{Youngjoon Jang},
  \bibinfo{person}{Dongsuk Oh}, {and} \bibinfo{person}{Heui-Seok Lim}.}
  \bibinfo{year}{2026}\natexlab{}.
\newblock \showarticletitle{CLEAR: Cross-Lingual Enhancement in Retrieval via
  Reverse-training}. In \bibinfo{booktitle}{\emph{Proceedings of the 64th
  Annual Meeting of the Association for Computational Linguistics (Volume 1:
  Long Papers)}}. \bibinfo{pages}{347--362}.
\newblock


\bibitem[Nawal et~al\mbox{.}(2024)]%
        {nawal2024effective}
\bibfield{author}{\bibinfo{person}{Noshin Nawal}, \bibinfo{person}{Sanju
  Basak}, {and} \bibinfo{person}{Rifat Shahriyar}.}
  \bibinfo{year}{2024}\natexlab{}.
\newblock \emph{\bibinfo{title}{Effective retrieval-augmented generation for
  open domain question answering in bengali}}.
\newblock \bibinfo{thesistype}{Ph.\,D. Dissertation}. \bibinfo{school}{Ph. D.
  dissertation}.
\newblock


\bibitem[Nzeyimana and Rubungo(2025)]%
        {nzeyimana2025kinyacolbert}
\bibfield{author}{\bibinfo{person}{Antoine Nzeyimana} {and}
  \bibinfo{person}{Andre~Niyongabo Rubungo}.} \bibinfo{year}{2025}\natexlab{}.
\newblock \showarticletitle{KinyaColBERT: A Lexically Grounded Retrieval Model
  for Low-Resource Retrieval-Augmented Generation}.
\newblock \bibinfo{journal}{\emph{arXiv preprint arXiv:2507.03241}}
  (\bibinfo{year}{2025}).
\newblock


\bibitem[Raihan and Zampieri(2025)]%
        {raihan2025tigerllm}
\bibfield{author}{\bibinfo{person}{Nishat Raihan} {and} \bibinfo{person}{Marcos
  Zampieri}.} \bibinfo{year}{2025}\natexlab{}.
\newblock \showarticletitle{TigerLLM-a family of Bangla large language models}.
  In \bibinfo{booktitle}{\emph{Proceedings of the 63rd Annual Meeting of the
  Association for Computational Linguistics (Volume 2: Short Papers)}}.
  \bibinfo{pages}{887--896}.
\newblock


\bibitem[Reza et~al\mbox{.}(2026)]%
        {reza2026krishokchat}
\bibfield{author}{\bibinfo{person}{Khan Raiyan~Ibne Reza},
  \bibinfo{person}{Sumaiya~Tabassum Nimi}, {and} \bibinfo{person}{Omar-Ibne
  Shahid}.} \bibinfo{year}{2026}\natexlab{}.
\newblock \showarticletitle{KrishokChat: A Provenance-Traceable Multi-Task
  Bengali Agricultural Benchmark with Safety-Critical Chemical Advisory}. In
  \bibinfo{booktitle}{\emph{Proceedings of the 18th Conference of the European
  Chapter of the Association for Computational Linguistics}}.
\newblock


\bibitem[Robertson and Zaragoza(2009)]%
        {robertson2009probabilistic}
\bibfield{author}{\bibinfo{person}{Stephen Robertson} {and}
  \bibinfo{person}{Hugo Zaragoza}.} \bibinfo{year}{2009}\natexlab{}.
\newblock \bibinfo{booktitle}{\emph{The probabilistic relevance framework: BM25
  and beyond}}. Vol.~\bibinfo{volume}{4}.
\newblock \bibinfo{publisher}{Now Publishers Inc}.
\newblock


\bibitem[Singh et~al\mbox{.}(2024)]%
        {singh2024farmerchat}
\bibfield{author}{\bibinfo{person}{Namita Singh}, \bibinfo{person}{Jacqueline
  Wang'ombe}, \bibinfo{person}{Nereah Okanga}, \bibinfo{person}{Tetyana
  Zelenska}, \bibinfo{person}{Jona Repishti}, \bibinfo{person}{Jayasankar G~K},
  \bibinfo{person}{Sanjeev Mishra}, \bibinfo{person}{Rajsekar Manokaran},
  \bibinfo{person}{Vineet Singh}, \bibinfo{person}{Mohammed~Irfan Rafiq},
  \bibinfo{person}{Rikin Gandhi}, {and} \bibinfo{person}{Akshay Nambi}.}
  \bibinfo{year}{2024}\natexlab{}.
\newblock \showarticletitle{Farmer.Chat: Scaling AI-Powered Agricultural
  Services for Smallholder Farmers}.
\newblock \bibinfo{journal}{\emph{arXiv preprint arXiv:2409.08916}}
  (\bibinfo{year}{2024}).
\newblock


\bibitem[Sorodoc et~al\mbox{.}(2025)]%
        {sorodoc2025garage}
\bibfield{author}{\bibinfo{person}{Ionut~Teodor Sorodoc},
  \bibinfo{person}{Leonardo~FR Ribeiro}, \bibinfo{person}{Rexhina Blloshmi},
  \bibinfo{person}{Christopher Davis}, {and} \bibinfo{person}{Adri{\`a} de
  Gispert}.} \bibinfo{year}{2025}\natexlab{}.
\newblock \showarticletitle{Garage: A benchmark with grounding annotations for
  rag evaluation}. In \bibinfo{booktitle}{\emph{Findings of the Association for
  Computational Linguistics: ACL 2025}}. \bibinfo{pages}{17030--17049}.
\newblock


\bibitem[Strich et~al\mbox{.}(2026)]%
        {strich2026t2}
\bibfield{author}{\bibinfo{person}{Jan Strich}, \bibinfo{person}{Enes~Kutay
  Isgorur}, \bibinfo{person}{Maximilian Trescher}, \bibinfo{person}{Chris
  Biemann}, {and} \bibinfo{person}{Martin Semmann}.}
  \bibinfo{year}{2026}\natexlab{}.
\newblock \showarticletitle{T2-RAGBench: Text-and-Table Aware
  Retrieval-Augmented Generation}. In \bibinfo{booktitle}{\emph{Proceedings of
  the 19th Conference of the European Chapter of the Association for
  Computational Linguistics (Volume 1: Long Papers)}}.
  \bibinfo{pages}{165--191}.
\newblock


\bibitem[Thakur et~al\mbox{.}(2021)]%
        {thakur2021beir}
\bibfield{author}{\bibinfo{person}{Nandan Thakur}, \bibinfo{person}{Nils
  Reimers}, \bibinfo{person}{Andreas R{\"u}ckl{\'e}}, \bibinfo{person}{Abhishek
  Srivastava}, {and} \bibinfo{person}{Iryna Gurevych}.}
  \bibinfo{year}{2021}\natexlab{}.
\newblock \showarticletitle{Beir: A heterogenous benchmark for zero-shot
  evaluation of information retrieval models}.
\newblock \bibinfo{journal}{\emph{arXiv preprint arXiv:2104.08663}}
  (\bibinfo{year}{2021}).
\newblock


\bibitem[Wang et~al\mbox{.}(2024)]%
        {wang2024multilinguale5}
\bibfield{author}{\bibinfo{person}{Liang Wang}, \bibinfo{person}{Nan Yang},
  \bibinfo{person}{Xiaolong Huang}, \bibinfo{person}{Linjun Yang},
  \bibinfo{person}{Rangan Majumder}, {and} \bibinfo{person}{Furu Wei}.}
  \bibinfo{year}{2024}\natexlab{}.
\newblock \showarticletitle{Multilingual E5 Text Embeddings: A Technical
  Report}.
\newblock \bibinfo{journal}{\emph{arXiv preprint arXiv:2402.05672}}
  (\bibinfo{year}{2024}).
\newblock


\bibitem[Wu et~al\mbox{.}(2024)]%
        {wu2024limits}
\bibfield{author}{\bibinfo{person}{Jie Wu}, \bibinfo{person}{Zhaochun Ren},
  {and} \bibinfo{person}{Suzan Verberne}.} \bibinfo{year}{2024}\natexlab{}.
\newblock \showarticletitle{What are the limits of cross-lingual dense passage
  retrieval for low-resource languages?}
\newblock \bibinfo{journal}{\emph{arXiv preprint arXiv:2408.11942}}
  (\bibinfo{year}{2024}).
\newblock


\bibitem[Yang et~al\mbox{.}(2024)]%
        {yang2024crag}
\bibfield{author}{\bibinfo{person}{Xiao Yang}, \bibinfo{person}{Kai Sun},
  \bibinfo{person}{Hao Xin}, \bibinfo{person}{Yushi Sun},
  \bibinfo{person}{Nikita Bhalla}, \bibinfo{person}{Xiangsen Chen},
  \bibinfo{person}{Sajal Choudhary}, \bibinfo{person}{Rongze~D Gui},
  \bibinfo{person}{Ziran~W Jiang}, \bibinfo{person}{Ziyu Jiang},
  {et~al\mbox{.}}} \bibinfo{year}{2024}\natexlab{}.
\newblock \showarticletitle{Crag-comprehensive rag benchmark}.
\newblock \bibinfo{journal}{\emph{Advances in Neural Information Processing
  Systems}}  \bibinfo{volume}{37} (\bibinfo{year}{2024}),
  \bibinfo{pages}{10470--10490}.
\newblock


\bibitem[Yeshambel et~al\mbox{.}(2025)]%
        {yeshambel2025dense}
\bibfield{author}{\bibinfo{person}{Tilahun Yeshambel}, \bibinfo{person}{Moncef
  Garouani}, \bibinfo{person}{Serge Molina}, {and} \bibinfo{person}{Josiane
  Mothe}.} \bibinfo{year}{2025}\natexlab{}.
\newblock \showarticletitle{Dense Retrieval for Low Resource languages-the Case
  of Amharic Language}. In \bibinfo{booktitle}{\emph{Proceedings of the 48th
  International ACM SIGIR Conference on Research and Development in Information
  Retrieval}}. \bibinfo{pages}{3098--3100}.
\newblock


\bibitem[Zeng and Tse(2006)]%
        {zeng2006exploring}
\bibfield{author}{\bibinfo{person}{Qing~T. Zeng} {and} \bibinfo{person}{Tony
  Tse}.} \bibinfo{year}{2006}\natexlab{}.
\newblock \showarticletitle{Exploring and developing consumer health
  vocabularies}.
\newblock \bibinfo{journal}{\emph{Journal of the American Medical Informatics
  Association}} \bibinfo{volume}{13}, \bibinfo{number}{1}
  (\bibinfo{year}{2006}), \bibinfo{pages}{24--29}.
\newblock


\bibitem[Zhang et~al\mbox{.}(2021)]%
        {zhang2021mr}
\bibfield{author}{\bibinfo{person}{Xinyu Zhang}, \bibinfo{person}{Xueguang Ma},
  \bibinfo{person}{Peng Shi}, {and} \bibinfo{person}{Jimmy Lin}.}
  \bibinfo{year}{2021}\natexlab{}.
\newblock \showarticletitle{Mr. TyDi: A multi-lingual benchmark for dense
  retrieval}. In \bibinfo{booktitle}{\emph{Proceedings of the 1st workshop on
  multilingual representation learning}}. \bibinfo{pages}{127--137}.
\newblock


\bibitem[Zhang et~al\mbox{.}(2023)]%
        {zhang2023miracl}
\bibfield{author}{\bibinfo{person}{Xinyu Zhang}, \bibinfo{person}{Nandan
  Thakur}, \bibinfo{person}{Odunayo Ogundepo}, \bibinfo{person}{Ehsan
  Kamalloo}, \bibinfo{person}{David Alfonso-Hermelo},
  \bibinfo{person}{Xiaoguang Li}, \bibinfo{person}{Qun Liu},
  \bibinfo{person}{Mehdi Rezagholizadeh}, {and} \bibinfo{person}{Jimmy Lin}.}
  \bibinfo{year}{2023}\natexlab{}.
\newblock \showarticletitle{Miracl: A multilingual retrieval dataset covering
  18 diverse languages}.
\newblock \bibinfo{journal}{\emph{Transactions of the Association for
  Computational Linguistics}}  \bibinfo{volume}{11} (\bibinfo{year}{2023}),
  \bibinfo{pages}{1114--1131}.
\newblock


\end{thebibliography}

\end{document}